\documentclass{salesforce_ai_research}

\usepackage{amsmath}
\usepackage{wrapfig}
\usepackage{xcolor}
\usepackage{colortbl}
\usepackage{makecell}
\usepackage{amssymb}
\usepackage{graphicx}
\usepackage{subfigure}
\usepackage{booktabs}
\usepackage{adjustbox}  
\usepackage{algorithm}
\usepackage{algpseudocode}
\DeclareMathOperator{\Quantile}{Quantile}
\usepackage{lipsum}
\usepackage{multirow}
\usepackage{shadowtext}
\usepackage{xspace}
\usepackage{listings}
\usepackage{fontawesome5}

\newcommand{\modelname}{Moirai 2.0}
\newcommand{\prevmodelname}{Moirai 1.0}
\newcommand{\hficon}[1][1.05em]{%
  \raisebox{-0.2\height}{\includegraphics[height=#1]{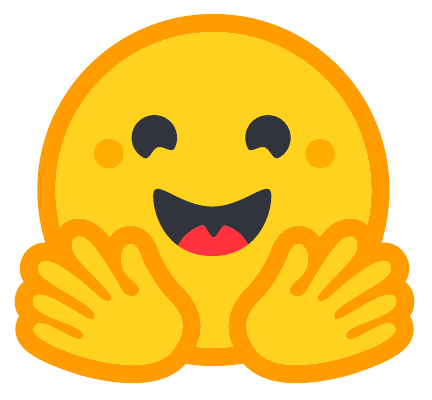}}%
}

\lstdefinestyle{pythonstyle}{
    language=Python,
    basicstyle=\ttfamily\small,
    keywordstyle=\color{blue}\bfseries,
    commentstyle=\color{gray}\itshape,
    stringstyle=\color{red},
    numberstyle=\tiny\color{gray},
    breaklines=true,
    breakatwhitespace=false,
    tabsize=4,
    showstringspaces=false,
    showspaces=false,
    showtabs=false,
    frame=tb,
    framerule=0.5pt,
    rulecolor=\color{gray!30},
    backgroundcolor=\color{gray!3},
    xleftmargin=0pt,
    xrightmargin=0pt,
    aboveskip=8pt,
    belowskip=8pt,
    captionpos=b,
    numbers=none,
    escapeinside={(*@}{@*)},
    morekeywords={async, await, def, return, if, for, in, None, True, False}
}

\title{Moirai 2.0: When Less Is More for Time Series Forecasting}
\author{
Chenghao Liu\thanks{Equal Contribution.}\quad Taha Aksu\footnotemark[1]\quad Juncheng Liu\footnotemark[1]\quad Xu Liu\footnotemark[1] \\ \textbf{Hanshu Yan\quad Quang Pham\quad Silvio Savarese\quad Doyen Sahoo\quad Caiming Xiong\quad Junnan Li\thanks{Corresponding author. Email: junnan.li@salesforce.com}}
\\
\textit{Salesforce AI Research} \\[0.5em]
\faGithub\, \url{https://github.com/SalesforceAIResearch/uni2ts} \\[0.5em]
\hficon\, \url{https://huggingface.co/Salesforce/moirai-2.0-R-small} 
}
\begin{document}
\maketitle

\begin{abstract}
We introduce \modelname{}, a decoder-only time-series foundation model trained on a new corpus of 36M series. The model adopts quantile forecasting and multi-token prediction, improving both probabilistic accuracy and inference efficiency. On the \textsc{Gift-Eval} benchmark, it ranks among the top pretrained models while achieving a strong trade-off between accuracy, speed, and model size.

Compared to \prevmodelname{}, \modelname{} replaces masked-encoder training, multi-patch inputs, and mixture-distribution outputs with a simpler decoder-only architecture, single patch, and quantile loss. Ablation studies isolate these changes---showing that the decoder-only backbone along with recursive multi-quantile decoding contribute most to the gains. Additional experiments show that \modelname{} outperforms larger models from the same family and exhibits robust domain-level results. In terms of efficiency and model size, \modelname{} is twice as fast and thirty times smaller than its prior best version, \prevmodelname{}-Large, while also performing better. Model performance plateaus with increasing parameter count and declines at longer horizons, motivating future work on data scaling and long-horizon modeling. We release code and evaluation details to support further research.
\end{abstract}






\section{Introduction}
Time series forecasting underpins capacity planning, anomaly response, and risk management across cloud infrastructure, observability, finance, energy, retail, and healthcare. 
Unlike text or images, time series exhibit nonstationarity, multi-scale temporal structure, irregular sampling, and incomplete observations, making generalization beyond a single domain both important and challenging. 
To tackle these, some early attempts on foundation models (FMs) for time series have been proposed and shown some  promising results on cross-domain reuse, zero/few-shot adaptation, and unified deployment at scale.

Moreover, foundation models for time series have rapidly evolved from early exploratory attempts just a few years ago to becoming a central research direction with significantly growing adoption in industry. Today, the landscape includes dozens of pretrained time series foundation models from both academia and industry. Representative examples from industry include TimesFM~\citep{das2024decoder} from Google, Chronos/Chronos-Bolt~\citep{ansari2024chronos} and Chronos 2~\citep{ansari2025chronos2} from Amazon, TOTO~\citep{cohen2024toto} from Datadog, and YingLong~\citep{wang2025output} from Alibaba. 
Community benchmarks from industry, such as \textsc{Gift-Eval} benchmark~\citep{aksu2024gift} and \textsc{fev-bench}~\citep{shchur2025fev}, also have accelerated progress by standarizing datasets, metrics and reporting performance of different models. 
A clear sign of this progress is that the \textsc{Gift-Eval} benchmark~\citep{aksu2024gift} has already received 25 foundation model submissions since its launch, showing the growing interest from the community. 

Our previous work, \prevmodelname{}~\citep{woo2024moirai}, was among the first attempts to scale the transformer architecture for time series. It demonstrated that large-scale time-series pretraining can yield strong generalization across domains, while also providing a blueprint for subsequent foundation models. While effective,~ \prevmodelname{} has also revealed several limitations. Its masked-encoder design led to inefficient data utilization during training, and the multi-patch setup constrained learning across different temporal frequencies. Moreover, while outputting a mixture of distributions was an intuitive way to enhance probabilistic forecasting, it proved empirically less effective in practice and added substantial complexity to both the model design and optimization process.

In this work, we introduce \modelname{}, a successor model that incorporates several architectural and training refinements. Key improvements include the adoption of quantile-based predictions with a refined loss function, a decoder-only autoregressive architecture, and multi-token prediction strategy with a single patch size, along with additional optimization techniques applied during both pretraining and inference. We have also curated a new pretraining datasets to train this model which includes $36M$ time series and $\sim295B$ observations. These changes substantially improve the accuracy and efficiency of our model, making it the best-performing model at the time of its release. As of today, it remains highly competitive, ranking 5th among 37 foundation models (including all size variants) on the \textsc{Gift-Eval} leaderboard, while offering one of the most favorable trade-offs between inference speed, model size, and accuracy.

The remainder of this paper is organized as follows. In~\Cref{sec:related}, we provide the background and discuss related work on time series foundation models.~\Cref{sec:method} introduces the architectural design and training strategies behind \modelname{}.
~\Cref{sec:training_data} lists the datasets we used to pretrain our new model.~\Cref{sec:experiments} presents evaluations on the \textsc{Gift-Eval} benchmark, including efficiency comparisons, scaling experiments, and ablation studies.~\Cref{sec:fworkandlimitations} outlines the limitations of our current approach and highlights future research directions. Finally,~\Cref{sec:conclusion} summarizes our findings and concludes the paper.

\section{Background \& Related Work}
\label{sec:related}
The time series point forecasting task is defined as predicting the future H values given a historical sequence with c values, \textit{i.e.} let the historical context be $Y_{1:c} = \{y_1, y_2, ..., y_c\}$ the target sequence to predict would be $Y_{c+1:c+H} = \{y_{c+1}, y_{c+2}, ... , y_{c+H}\}$. The forecasting task can naturally extend to the multivariate setting where one is expected to predict multiple variates simultaneously \textit{i.e.} $Y^1_{c+1:c+H} ,Y^2_{c+1:c+H}, ... Y^N_{c+1:c+H} $. \modelname{}, however, does not support cross-variate forecasting. Instead, we treat multivariate forecasting as a collection of independent univariate tasks. \modelname{} does not only predict point but quantile forecast, where the difference in definition comes at the prediction level. For quantile forecast given the same historical context $Y_{1:c}$, the target prediction is now the quantile levels for each future time step $Q_{c+1:c+H} = \{Q_{c+1}, Q_{c+2}, ..., Q_{c+H}\}$ where $Q_i = \{q_{l_1}^i,q_{l_2}^i,...,q_{l_q}^i, \}$, $l_1$ through $l_q$ being the quantile levels. We set the quantile levels as $\{0.1, 0.2, ..., 0.9\}$. Quantile forecasts offer a principled way to model uncertainty, as they are directly optimized for probabilistic forecasting accuracy. 

The family of time series forecasting models can be categorized into 3 classes: statistical models, deep-learning models and foundation models. The statistical models work locally and rely solely on historical data statistics to predict future values. Some of the most widely used ones are ETS~\citep{hyndman2008forecasting}, Theta~\citep{garza2022statsforecast}, and ARIMA~\citep{box1970distribution}.

Deep learning models, in contrast, are often dataset-specific, as each dataset typically requires training a dedicated model. Examples include N-BEATS~\citep{Oreshkin2019NBEATSNB}, DLinear~\citep{Zeng2022AreTE}, and DeepAR~\citep{Flunkert2017DeepARPF}, which are based on pre-transformer architectures. In contrast, transformer-based approaches such as Autoformer~\citep{Wu2021AutoformerDT}, Crossformer~\citep{zhang2023crossformer}, and PatchTST~\citep{nietime} have also been proposed.

In the last few years, the success of foundation models in language and vision has also attracted time series researchers. What initially started as a handful of early attempts~\citep{Goswami2024MOMENTAF, rasul2023lag, woo2024moirai} has now grown into a rapidly expanding landscape, with more than 25 time series foundation models being publicized in the last 2 years. 

The architecture and training strategies of time series foundation models vary considerably. Notably, two prominent approaches do not directly rely on transformer architectures: TabPFN-TS~\citep{hoo2025tables}, which employs a Prior-Data Fitted Network (PFN), and TiRex~\citep{auer2025tirex}, which is built on an xLSTM. The majority of other models, however, rely on transformer variants. Within the transformer family, some models adopt an encoder-only design, such as Chronos-2~\citep{ansari2025chronos2}, Yinglong~\citep{wang2025output}, and the first version of our Moirai~\citep{woo2024moirai}. In this setup, a forecasting head is placed on top of the encoder backbone. This approach can mitigate error accumulation and accelerate inference, particularly when autoregression is avoided. In contrast, another group follows the decoder-only paradigm inspired by large language models (LLMs). Examples include Moirai-MoE~\citep{liu2024moirai}, TimesFM family of models~\citep{das2024decoder}, and Sundial~\citep{liu2025sundial}. Others adopt a hybrid encoder–decoder architecture, such as Kairos~\citep{feng2025kairos}, Chronos, Chronos-Bolt~\citep{ansari2024chronos}, and FlowState~\citep{graf2025flowstate}. A further axis of differentiation is the choice of output representation and loss function. Some models predict point forecasts directly~\citep{das2024decoder,feng2025kairos}. Others, like Moirai, predict full distributions~\citep{woo2024moirai}, or employ flow-based losses to learn continuous distributions from which forecasts can be sampled as applied by ~\citet{graf2025flowstate,liu2025sundial}, or directly output quantiles~\citep{ansari2025chronos2,auer2025tirex,wang2025output, ansari2024chronos}.

\modelname{} adopts a decoder-only architecture, with newly incorporated training and inference strategies to limit error accumulation and inference slowdown. Unlike point or distribution-based approaches, it outputs quantile forecasts, directly aligned with the CRPS metric through optimization with the quantile (pinball) loss.




\section{Model Architecture and Training}
\label{sec:method}
\begin{figure}[th]
    \centering
    \includegraphics[width=0.9\linewidth]{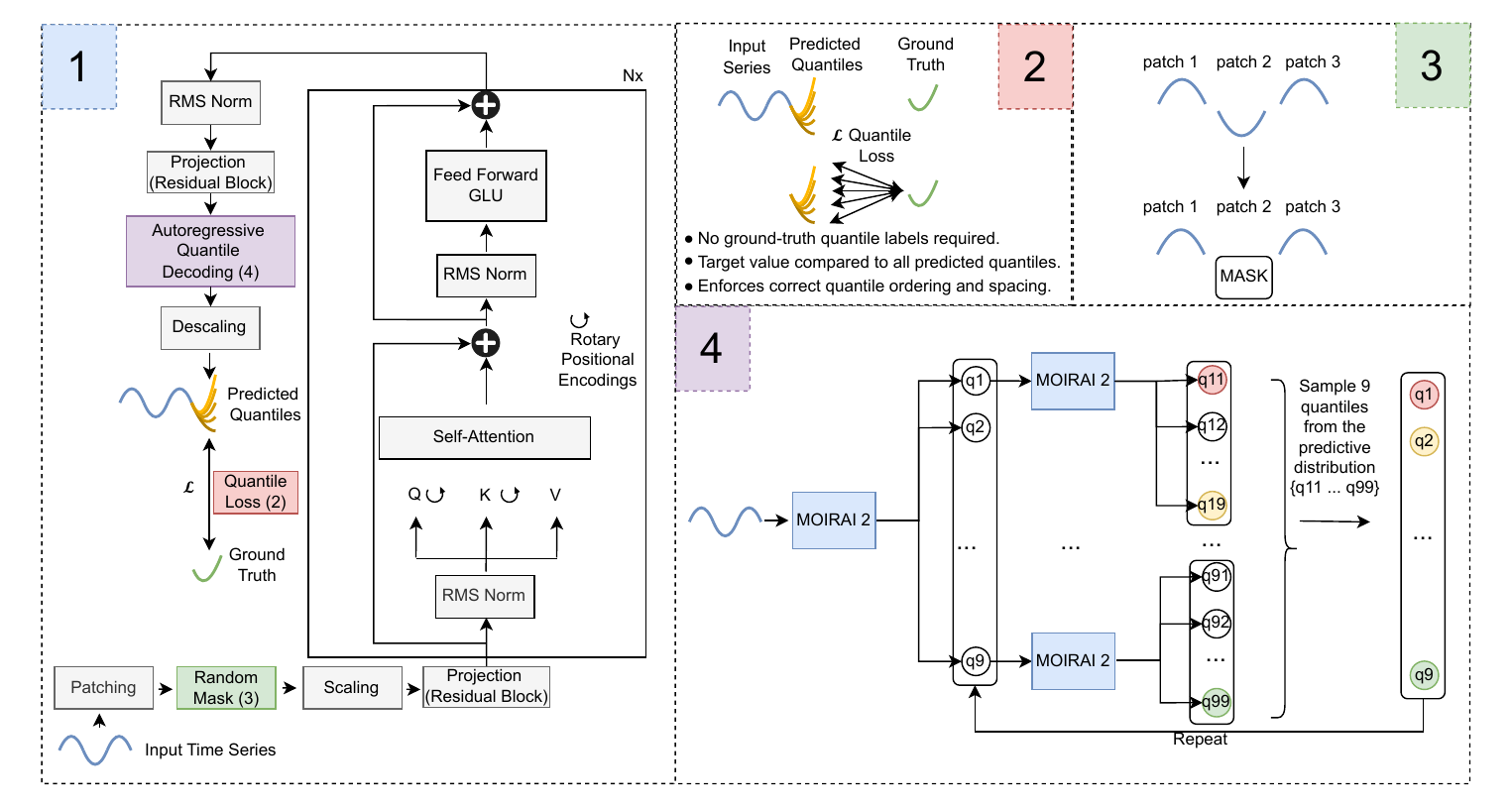}
    \caption{Overview of the \modelname{} architecture. Panel~1 illustrates the end-to-end pipeline from patched input time series through the transformer backbone to predicted quantiles. Panel~2 highlights the quantile loss, which compares each ground-truth value against all predicted quantiles without requiring quantile labels, enforcing correct ordering and spacing. Panel~3 depicts patch-level random masking used to improve robustness. Panel~4 shows a simplified view of the autoregressive multi-step quantile decoding strategy, where quantile forecasts are recursively rolled out to construct predictive distributions across the horizon.}
    \label{fig:mainfig}
\end{figure}
In this section, we first describe the model architecture of \modelname{} in ~\Cref{sec:model_arch} and present the details of training and inference in~\Cref{sec:training_inference}. 
Finally, in~\Cref{Sec:comparison_moirai_1_2}, we discuss and summarize the differences between \modelname{} and Moirai 1.0.

\subsection{Model Architecture}
\label{sec:model_arch}
For efficient training, \modelname{} is structured as a decoder-only model built upon the Transformer architecture. We provide an overview of our architecture in~\Cref{fig:mainfig}. 

\paragraph{Input Projection} The input layers are responsible for preprocessing the raw time-series into input tokens for the transformer architecture. Currently, \modelname{} is designed for univariate time-series input. Multivariate data is accommodated by treating each variable as an independent univariate series, an approach often found to perform well \cite{nietime}. Instance normalization \cite{kim2021reversible} is applied to each time series individually to ensure the model is robust to diverse and non-stationary input scales. Adhering to established practices \cite{woo2024moirai}, the input time-series is first partitioned into contiguous, non-overlapping $T$ patches. Missing values are explicitly handled by concatenating a corresponding binary indicator $\mathbf{m}_i$ (representing presence or absence) with the value at each time step within the patch. Each resulting patch, augmented with this missing value information as $\hat {\mathbf{x}}_i = \mathbf{x}_i || \mathbf{m}_i$,  is then processed into an input token with a residual block \cite{he2016deep}. Formally, given an input patch size of $p_{in}$ and token embedding dimension $d$, this input projection with a residual block defines a mapping $\mathbb{R}^{2p_{in}} \rightarrow \mathbb{R}^{d}$ as shown below: 

\begin{equation}
   \mathbf{z}_i = \text{PatchEmbed}(\hat {\mathbf{x}}_i) = \text{SiLU}(\mathbf{W}(\hat {\mathbf{x}}_i) + \mathbf{b}) + \hat {\mathbf{x}}_i \in \mathbb{R}^d, i = 1, ... T, 
\end{equation}
where $\mathbf{W}$ and $\mathbf{b}$ are weights and bias for the input projection. 

Additionally, we also apply normalization before input projection. Notably, as mentioned in~\citet{liu2024moirai}, applying instance normalization globally risks future information leakage in decoder-only models. We aim to avoids this by computing normalization statistics solely from the first 30$\%$ of the time series, reserving the subsequent 70$\%$ segment for the causal pre-training task.

\paragraph{Stacked Transformer} Following the input projection, the sequence of tokens is processed by multiple stacked Transformer layers, \textit{c.f.}Panel $1$ in~\Cref{fig:mainfig}. Each layer within this stack adheres to the standard Transformer architecture \cite{vaswani2017attention}, primarily comprising two sub-layers: a causal multi-head self-attention mechanism followed by a feed-forward network. The causal nature of the self-attention ensures that the prediction for a given token depends only on valid preceding tokens and the current token itself, maintaining the autoregressive property required by the decoder-only framework. Residual connections are employed around each of the two sub-layers, followed by layer normalization, promoting stable training and effective gradient flow through the deep network. The output sequence from the final Transformer layer represents the learned contextual embeddings of the input patches.

\paragraph{Output Projection}
The final stage involves projecting the sequence of processed token embeddings, derived from the stacked transformer layers, onto the target prediction space. An output residual block \cite{he2016deep} performs this transformation, converting each $d$-dimensional token embedding into forecasts. Two key features are incorporated into this projection: First, to facilitate probabilistic predictions, the output dimension is scaled to accommodate $n_q$ quantile estimates per time step \cite{wen2017multi}. Second, the model employs \textbf{multi-token prediction}, generating forecasts for $n_{token}$ future patches from each output token representation to improve effectiveness and efficiency for long term forecast. Thus, for an output patch size $p$, the output projection layer effectively realizes a mapping $\mathbb{R}^{d} \rightarrow \mathbb{R}^{n_{token} \times n_q \times p}$, transforming the learned representations into structured quantile forecasts for multiple future patches.

We choose multi-token prediction instead of single-token prediction to improve efficiency, especially for long-term forecasting with relatively large prediction lengths. 
Additionally, multi-token prediction also helps to reduce the error accumulation in long-term forecasting.

\paragraph{Loss Function}
To support probabilistic forecasting capabilities, \modelname{} is trained by optimizing the quantile loss (also known as the pinball loss). This objective function enables the model to learn the conditional distribution of future time series values. Specifically, the model is configured to predict $n_q=9$ distinct quantile levels, equidistantly spaced from $0.1$ to $0.9$ ($Q = \{0.1, 0.2, \dots, 0.9\}$). The quantile loss for a single prediction at time step $t$, for a given quantile level $q \in Q$, is defined as:
$$
\ell_{q}(y_t, \hat{y}_t^{(q)}) =
\begin{cases}
q (y_t - \hat{y}_t^{(q)}) & \text{if } y_t \ge \hat{y}_t^{(q)}, \\
(1 - q) (\hat{y}_t^{(q)} - y_t) & \text{if } y_t < \hat{y}_t^{(q)},
\end{cases}
$$
where $y_t$ is the ground truth value at time $t$, and $\hat{y}_t^q$ is the model's predicted value for the $q$-th quantile at that time step. During training, the total loss is calculated by averaging the quantile loss across all predicted quantile levels ($q \in Q$) and all $H$ time steps in $K$ predicted patches, while ignoring contributions from any masked or missing target values. We formalize the total loss as follows:
\newcommand{\Hlen}{H} 
\newcommand{\Qset}{Q}
\newcommand{\Qnum}{|Q|}
\begin{align}
\mathcal{L}_{\text{Q}} 
& = \frac{1}{\Hlen\,\Qnum}
\sum_{t=1}^{\Hlen}\sum_{q \in\Qset} \ell_q(y_t, \hat{y}_t^{(q)}) \\ 
 & = \frac{1}{\Hlen\,\Qnum}
\sum_{t=1}^{\Hlen}\sum_{q \in\Qset}
\left[
q\,\max\big(y_{t}-\hat{y}^{(q)}_{t},\,0\big)
+ (1-q)\,\max\big(\hat{y}^{(q)}_{t}-y_{t},\,0\big)
\right],
\quad \Hlen=Kp,
\end{align}
Note that, by default , we treat all quantile levels equally, which if often preferable in practice and encourages accurate and well-calibrated quantile forecasts.
However, when specific regions of the distribution are of greater interest, we can assign different weights (e.g., $w_q$) on selected quantiles to emphasize them. 
We discuss more on this in the last paragraph of ~\Cref{Sec:comparison_moirai_1_2}

\subsection{Training and Inference} 
\label{sec:training_inference}
\paragraph{Training} 

To enhance model robustness, \textbf{patch-level random masking} is utilized: $50\%$ of input patches in each training sample are randomly masked before being processed by the model. This encourages robust representation learning and improves handling of missing data segments.
Since normalization statistics are computed using only the initial $30\%$ segment of each sample, distribution shifts between this segment and the remaining $70\%$ utilized for training may arise, potentially causing significant instability during the training process. To mitigate this, we employ z-score based anomaly detection, samples are filtered out if the latter $70\%$ segment exhibits significant statistical deviation from the initial segment used for normalization. Following a similar training procedure as Moirai 1.0 \cite{woo2024moirai}, \modelname{} is trained for $100,000$ steps with a batch size of $256$. We utilize the AdamW optimizer  configured with a learning rate of $1 \times 10^{-3}$, weight decay of $1 \times 10^{-1}$, $\beta_1=0.9$, and $\beta_2=0.98$. The learning rate schedule includes a linear warmup over the first 10,000 steps, followed by cosine annealing. Training is conducted using bf16 mixed-precision arithmetic to enhance efficiency.


\paragraph{Inference}

In standard autoregressive decoding, a model generates forecasts step by step by feeding its output back as input. This works naturally when the model predicts a single value per step, since the input and output dimensions match. 

For \emph{quantile forecasting}, however, the model outputs multiple values (quantiles) at each step. Feeding all quantiles directly back into the model creates a dimensional mismatch, and collapsing them to a single value (e.g., the median or mean~\citep{das2024decoder}) discards important information about uncertainty. 

To address this, we propose to use \textbf{autoregressive multi-quantile decoding}. Conceptually, this procedure resembles a \emph{beam search with depth~2}.
Given the forecast with multiple quantiles at the previous step $t-1$, instead of committing to a single quantile path with the median value, we temporarily expand the search tree at the current step $t$, then collapse it back to a fixed set of quantiles. 
At each decoding step $t$, we first \emph{expand} the forecast with several quantiles from step $t-1$ into a larger number of candidate quantiles (e.g., $9\times9=81$). Then, we \emph{collapse} these samples back into the desired quantile set (e.g., 9 levels) by sampling from the expanded set and use the collapsed quantiles as the forecast for step $t$. This procedure preserves forecast uncertainty while keeping the inference tractable.
\begin{algorithm}[h]
\caption{Autoregressive Multi-Quantile Decoding with Initialization (depth-2 expand $\rightarrow$ collapse)}
\label{alg:multi_quantile}
\begin{algorithmic}[1]
\Require Context $Y_{1:c}$; target quantile set $\mathcal{Q}$ with $m{=}|\mathcal{Q}|$; horizon $H$
\Statex \textbf{Initialization (first prediction, no expansion).}
\State Predict the first-step quantiles directly from the context:
      $\{\hat{y}_{c+1}^{(q)}\}_{q\in\mathcal{Q}} \gets \textsc{Moirai2}(Y_{1:c})$.
\State Append $\{\hat{y}_{c+1}^{(q)} : q\in\mathcal{Q}\}$ to the context.

\Statex \textbf{Autoregressive steps (beam-search depth 2).}
\For{$t = c+1$ to $c+H-1$}
    \State \textbf{Expand:} form $m$ histories by appending each $\hat{y}_{t}^{(q_1)}$ to the current context.
    \State From each expanded history, decode one step ahead to obtain
          $\{\hat{y}_{t+1}^{(q_1,q_2)}\}_{q_2\in\mathcal{Q}}$; pool all candidates
          \[
            \mathcal{S}_{t+1} \gets \bigl\{\hat{y}_{t+1}^{(q_1,q_2)} : q_1,q_2 \in \mathcal{Q}\bigr\},
            \quad |\mathcal{S}_{t+1}| = m^2 .
          \]
    \State \textbf{Collapse:} for each $q \in \mathcal{Q}$, set
          \[
            \hat{y}_{t+1}^{(q)} \gets \Quantile_{q}\!\left(\mathcal{S}_{t+1}\right).
          \]
    \State Append $\{\hat{y}_{t+1}^{(q)} : q\in\mathcal{Q}\}$ to the context.
\EndFor
\Statex \emph{Note (patch-level decoding).} For clarity the pseudocode shows single-value prediction; in practice, the model outputs multiple patch tokens per iteration.
\end{algorithmic}
\end{algorithm}

In practice, this approach allows us to maintain the benefits of quantile forecasting while avoiding dimensional mismatch. It can be viewed as sampling multiple plausible futures (expansion) and then summarizing them into a coherent set of quantiles  (collapse).

\subsection{Discussion and Comparison on \modelname{} and Moirai 1.0}
\label{Sec:comparison_moirai_1_2}
In this section, we provide some discussions and comparisons on architecture changes from Moirai 1.0 to \modelname{}.
From the architecture perspective, we mainly have 3 changes: 1) use decoder-only for \modelname{} instead of masked encoder in Moirai 1.0, 
2) change to single-path size in \modelname{} from multi-patch size in Moirai 1.0, 
3) formulate the training loss as quantile loss in \modelname{} instead of distribution loss Moirai 1.0. 

\paragraph{Decoder-only in \modelname{} vs Masked encoder in Moirai 1.0}

\prevmodelname{}~\citep{woo2024moirai} adopts a masked-encoder architecture, where both context and prediction lengths are randomly sampled during training to support flexible downstream usage. However, this design yields only a single loss per sampled input configuration, leading to suboptimal data utilization and slower training efficiency. In contrast, \modelname{} replaces the masked encoder with a decoder-only architecture, enabling more effective use of training data and simplifying the overall modeling pipeline.

In terms of data efficiency during training, \modelname{}, built upon causal decoder-only transformers, directly computes $T{-}1$ losses across all tokens or patches for a time series with $T$ tokens. In contrast, the masked-encoder architecture used in Moirai-1.0, assuming a masking rate of 15\%, engages only 15\% of tokens in the loss computation, producing a single loss value for each specific pair of context and prediction lengths. This difference in training strategy makes \modelname{} substantially more data-efficient during training.

Another advantage of using the decoder-only autoregressive architecture is the potential inference speedup with KV caches \citep{pope2023efficiently}. A KV cache stores the intermediate key–value attention representations computed from the input tokens during the first forward pass; at later decoding steps, the model can reuse these cached representations instead of recomputing them, reducing inference cost.
Thus, when the model is asked to provide predictions repeatedly (e.g., n times), autoregressive models such as \modelname{} can pre-fill the context once and reuse cached KV states, avoiding redundant computation. In contrast, masked-encoder models such as \prevmodelname{} must recompute key–value features n times from scratch, making them less efficient for repeated queries.
This scenario is not uncommon in real-world forecasting, where end users often do not know the exact prediction horizon in advance—they may start by forecasting a few steps and then gradually extend the prediction as needed. To quantify the potential benefit, we conducted a case study by implementing a KV-cache–enabled version of \modelname{}. The results show that inference speedup scales with both context and prediction length: for a context of approximately 10K and a prediction length of 1K, KV caching provides up to a $4\times$ speedup, while extending the prediction length to 10K increases the gain to $17\times$.




\paragraph{Multi patch size vs Single patch size}
In \modelname{}, we standardize to a single patch size, in contrast to the multi-patch design of \prevmodelname{}. Empirically, this simplification improves both computational efficiency and forecasting accuracy, while also simplifying the implementation of training and inference stages.

\paragraph{Quantile forecast vs Distribution forecast}
In Moirai 1.0, the model outputs the distribution parameters for a mixture of distributions,and generates forecasts by sampling from this mixture. During inference, 100 samples per step are typically drawn, and during training the distribution negative log likelihood (NLL) loss is used. 
In contrast, we use quantile loss for training \modelname{}, which generates $|Q|$ quantiles as the forecast. 
The quantile loss can directly optimize operational quantiles (e.g., using $q=0.9$ for capacity planning) and it naturally handles asymmetric penalties (over- vs under-prediction). 
Compared to the distribution NLL loss, which may suffer from variance collapse, explosion, or unstable gradients under outliers, the quantile loss is more robust to extreme tails.

\section{Pretraining Datasets}
\label{sec:training_data}
Our pretraining corpus is constructed from five complementary sources: 
(1) the non-leaking \textsc{Gift-Eval Pretrain} dataset, (2)  train split of \textsc{Gift-Eval TrainTest} dataset~\citep{aksu2024gift}
(3) additional series generated via Chronos-Mixup, 
(4) KernelSynth data~\citep{ansari2024chronos}, 
and (5) anonymized internal Salesforce CloudOps telemetry data. In combination these five components add up to $36M$ time series, with $\sim295B$ observations, providing a diverse mixture of real-world and synthetic time series, covering a wide range of domains, frequencies, and temporal characteristics. We describe each component in more detail below.

\paragraph{Gift-Eval Pretrain and Gift-Eval TrainTest}
The original Moirai model, reported in the GIFT-Eval leaderboard, was trained using a non-leaking version of the pretraining corpus \textsc{Gift-Eval Pretrain}~\citep{aksu2024gift}. Our new pretraining corpus also includes this same non-leaking dataset as its foundation. It has ~$3.25$M time series with a total of ~$230$B observations\footnote{Note that GIFT-Eval Pretrain is a subset of LOTSA introduced by~\citet{woo2024moirai}. While the original paper reports the dataset size as 27B observations, this discrepancy arises solely from differences in counting: they treat each multivariate series as a single observation, whereas in our accounting each variable is counted independently.}. \textsc{Gift-Eval Pretrain} provides a large and diverse set of time series carefully curated to avoid overlap with benchmark evaluation tasks. Additionally we add the train split of \textsc{Gift-Eval TrainTest} dataset to train \modelname{}, which includes $144$K time series. For details about both datasets composition and construction methodology, we refer the reader to~\citet{aksu2024gift}.

\paragraph{Chronos-Mixup}



Time Series Mixup (TSMixup) is proposed originally by~\citet{ansari2024chronos} which takes inspiration from original Mixup used in image classification~\citep{zhang2017mixup}. It randomly samples $k \sim \mathcal{U}\{1, K\}$ time series of length $l \sim \mathcal{U}\{l_{min}, l_{max}\}$ from a distribution of training data then takes their convex combination.
We follow~\citet{auer2025tirex}, and generate significantly more and longer series and following their configuration we set the max value of $K=4$, and set $L_{min}=128$, $L_{max}=4096$. We only use non leaking subset of datasets from Chronos data to generate the mixup to keep our training free of leakage. We generate $30$M time series that add up to $63$B observations in total.
\paragraph{KernelSynth}
We also include the KernelSynth data provided by~\citet{ansari2024chronos} which generate synthetic time series using Gaussian processes. Their method constructs a kernel bank including kernels for trend, local change and seasonalities. The final kernel time series is constructed by sampling kernel from the bank and combining them with random binary operations $+$ or $\times$. This set contains $1$M time series, with $1.02$B observation in total.

\paragraph{Internal Salesforce data}
A core component of our pre-training corpus is derived from internal salesforce telemetry data. This dataset comprises approximately 2.15M univariate time series, totaling roughly 1.48B observations at a daily granularity. The data covers a period of approximately one year, starting from January 2024. To ensure data quality, time series containing substantial amounts of missing values were filtered out prior to training.





\begin{figure} [!htb]
    \centering
    \includegraphics[width=1\linewidth]{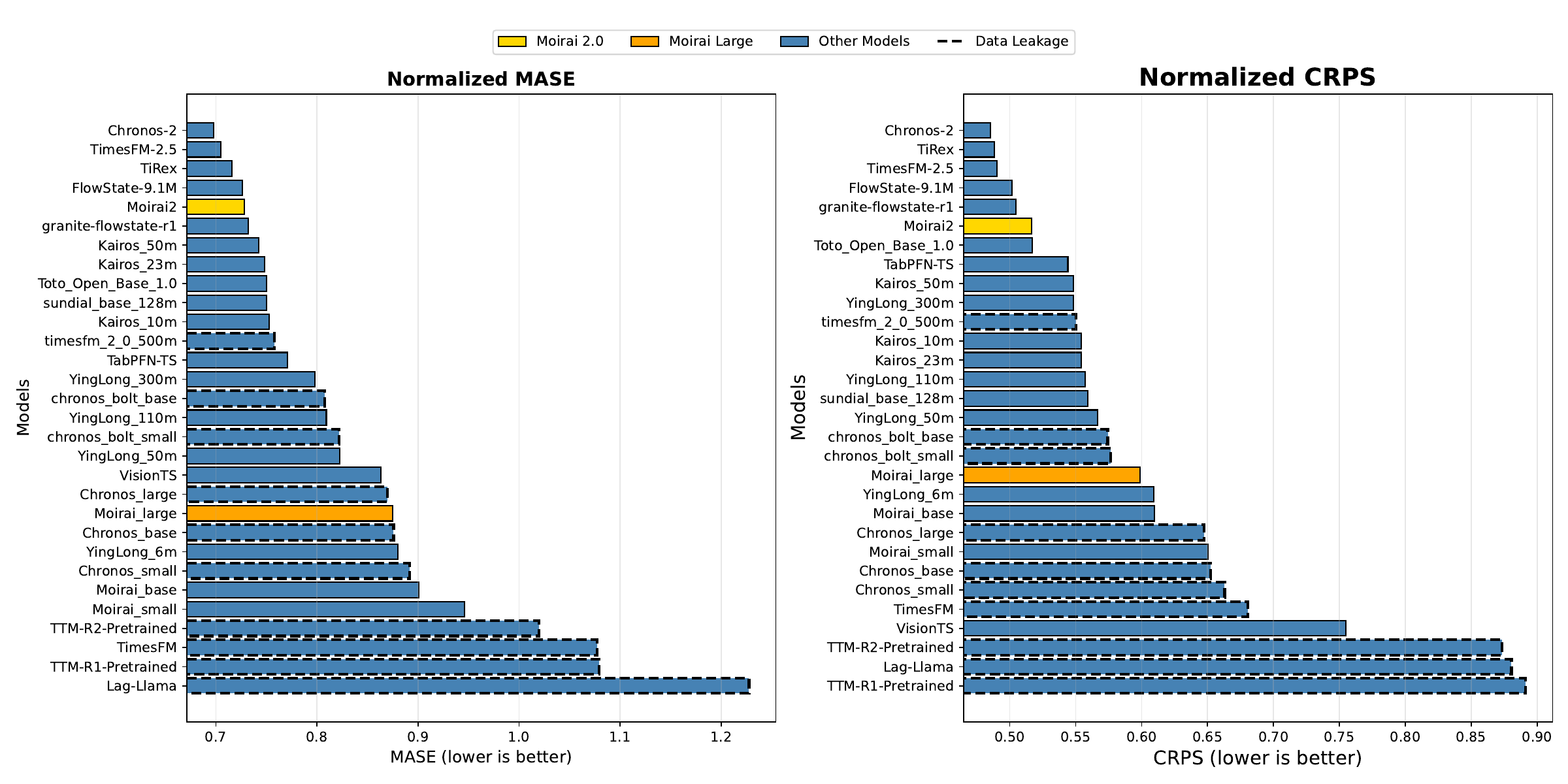}
    \caption{\textsc{GiftEval} benchmark results for pretrained, zero-shot foundation models. Ensemble methods and models lacking reproducible code are excluded. Bars show normalized MASE (left) and normalized CRPS (right), where lower is better. \modelname{} and its large variant rank among the top models under both metrics.}
    \label{fig:gifteval_results}
\end{figure}

\begin{figure}
    \centering
    \includegraphics[width=1\linewidth]{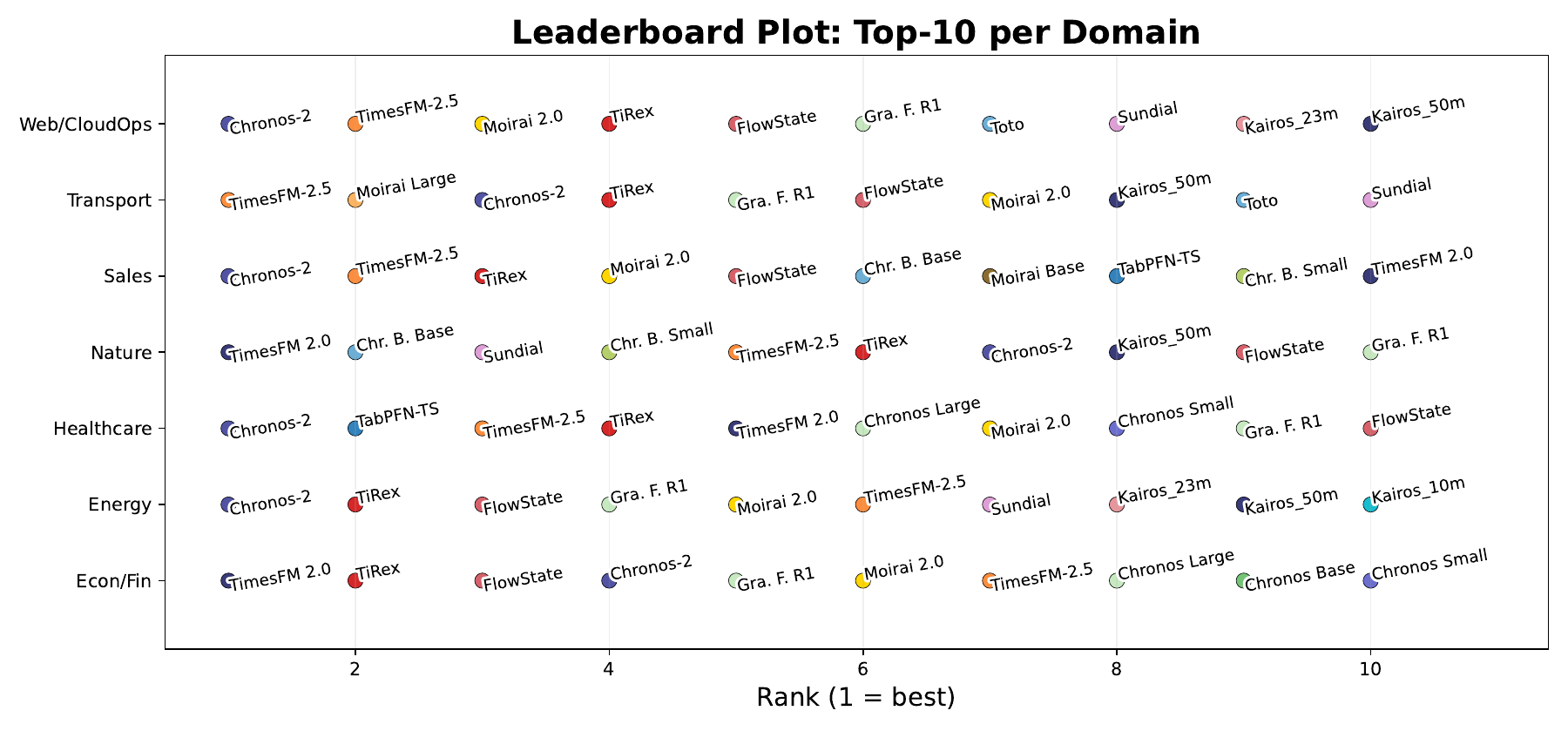}
    \caption{\textsc{GiftEval} leaderboard results broken down by domain. For each domain, we display the top-10 foundation models ordered by their MASE rank (lower is better).}
    \label{fig:leaderboard-per-domain}
\end{figure}

\section{Evaluation}
\label{sec:experiments}

In this section, we present experimental results that demonstrate the capabilities of our model compared to other state-of-the-art foundation models. We primarily evaluate on the GIFT-Eval benchmark~\citep{aksu2024gift}, comparing against all available pretrained foundation models with reproducible code and no test-data leakage. In addition to overall benchmark performance, we analyze \modelname{} from several perspectives: inference speed and models size and their relationship with accuracy compared to top leaderboard models, and the effect of scaling \modelname{} to larger parameter counts. We also report domain-specific results across models and conclude with an ablation study to quantify the contribution of individual components.
\begin{figure}
    \centering
    \includegraphics[width=1\linewidth]{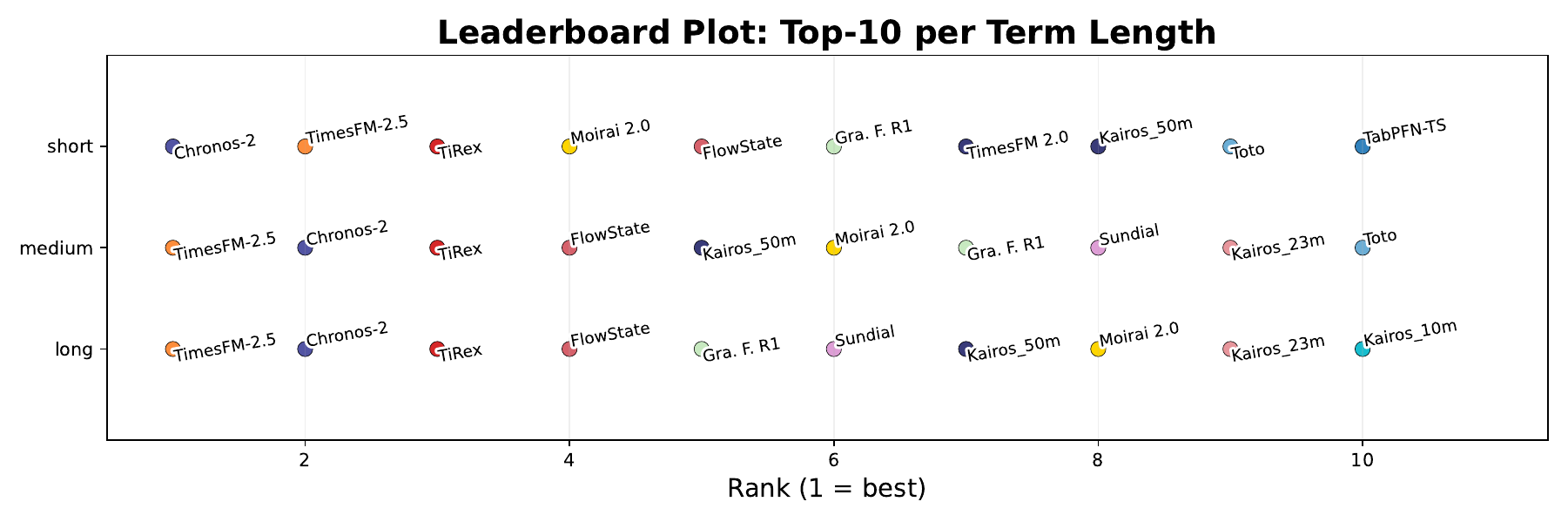}
    \caption{\textsc{GiftEval} leaderboard results broken down by prediction length. For each domain, we display the top-10 foundation models ordered by their MASE rank (lower is better).}
    \label{fig:leaderboard-per-term-length}
\end{figure}

\subsection{Results on GIFT-Eval Benchmark}
We evaluate \modelname{} on the comprehensive GIFT-Eval benchmark, comparing it against state-of-the-art foundation models listed on the leaderboard. GIFT-Eval covers 97 task configurations across 55 datasets, spanning diverse domains, frequencies, and prediction lengths. For this study, we exclude agentic solutions, fine-tuned models, and foundation models without publicly available replication code. This filtering leaves 30 foundation models for comparison, including \texttt{Chronos-2}~\citep{ansari2025chronos2}, \texttt{TimesFM-2.5}, \texttt{TimesFM-2.0}, \texttt{TimesFM}~\citep{das2024decoder}, \texttt{TiRex}~\citep{auer2025tirex}, \texttt{FlowState} and \texttt{Granite-FlowState-R1}~\citep{graf2025flowstate}, \texttt{Kairos (10/23/50M)}~\citep{feng2025kairos}, \texttt{Toto}~\citep{cohen2024toto}, \texttt{Sundial}~\citep{liu2025sundial}, \texttt{TabPFN-TS}~\citep{hoo2025tables}, \texttt{YingLong (6/50/110/300M)}~\citep{wang2025output}, \texttt{Chronos (small/base/large)} and \texttt{Chronos-Bolt (small/base)}~\citep{ansari2024chronos}, \texttt{Moirai (small/base/large)}~\citep{woo2024moirai}, \texttt{TTM-R1-Pretrained}~\citep{ekambaram2024tiny}, and \texttt{Lag-Llama}~\citep{rasul2023lag}.

Following leaderboard protocols, we report normalized MASE and CRPS, where each model’s score is divided by the seasonal-naive baseline and aggregated using the geometric mean across tasks. As shown in \Cref{fig:gifteval_results}, \modelname{} attains the 5th and 6th positions on MASE and CRPS, respectively—substantially outperforming its predecessor Moirai-Large despite having fewer parameters.

\paragraph{Fine Grained Results}
Beyond overall benchmark scores, we analyze \modelname{}’s performance across domains in GIFT-Eval. As shown in \Cref{fig:leaderboard-per-domain}, \modelname{} consistently appears in the top-10 across most domains, with the notable exception of Nature tasks. This gap may suggest that the pretraining corpus underrepresents natural and environmental time series, motivating future work to balance domain coverage. Compared to its predecessor Moirai-Large, \modelname{} delivers stronger results in nearly all domains, with Transport being the only case where Moirai-Large remains highly competitive. An additional perspective emerges when grouping tasks by forecast horizon~\Cref{fig:leaderboard-per-term-length}: \modelname{} ranks 4th, 6th, and 8th on short, medium, and long prediction lengths, respectively. This trend highlights that while the model is highly competitive for short-range forecasting, its relative advantage diminishes as the prediction horizon grows.


\begin{figure}[!htb]
    \centering
    \includegraphics[width=1\linewidth]{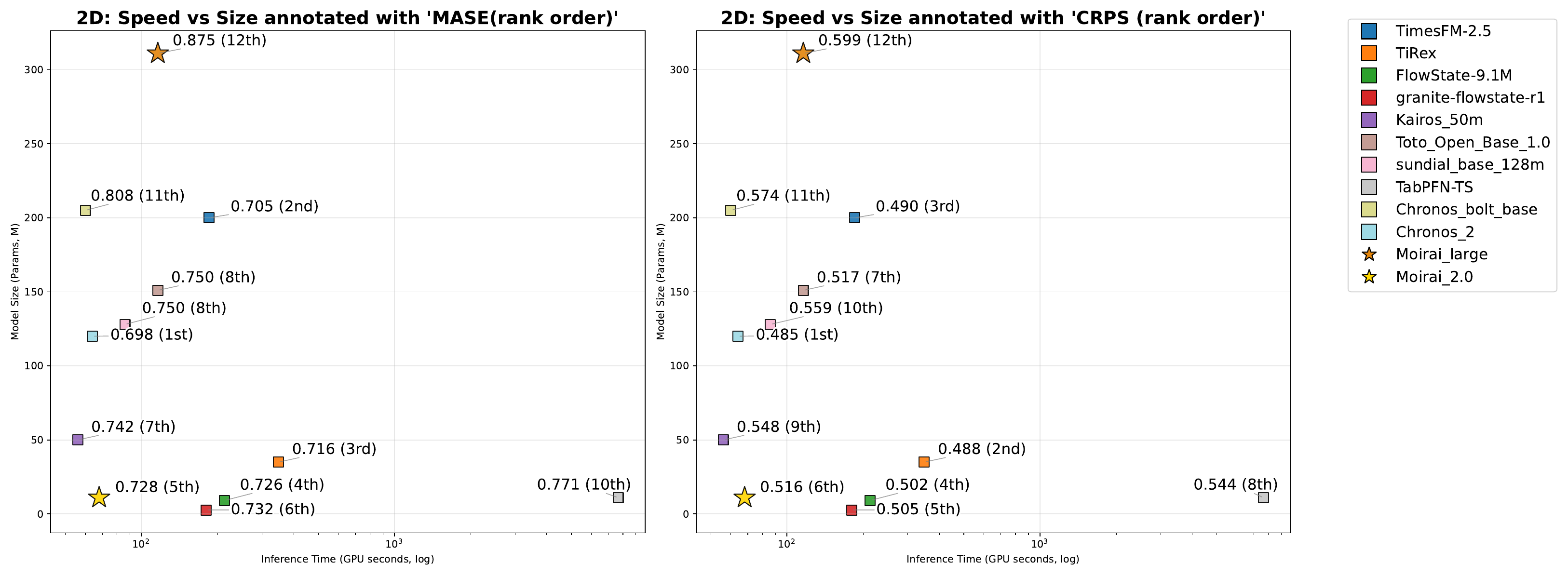}
    \caption{Speed--parameter count comparison across foundation models. Each point shows inference time (x-axis) vs.\ model size (y-axis), annotated with performance rank. The left plot annotates model rank by MASE, while the right plot by CRPS (lower is better). \modelname{} variants achieve competitive accuracy while offering favorable size and inference efficiency.
}
    \label{fig:perf_size_speed}
\end{figure}

\subsection{Efficiency Comparison}
To assess inference efficiency, we conducted experiments on 12 representative tasks from GIFT-Eval, selected to cover diverse dataset–frequency combinations and forecast horizons. Specifically, we included short-horizon tasks such as \texttt{M4--yearly}, \texttt{M4--hourly}, \texttt{Hospital--monthly}, \texttt{US Births--weekly}, \texttt{Electricity--daily}, \texttt{Saugeenday--daily}, \texttt{Bizitobs\_l2c--hourly}, \texttt{SZ\_Taxi--15 minutely}, and \texttt{Solar--hourly}, as well as medium/long-horizon tasks including \texttt{SZ\_Taxi--15 minutely}, \texttt{Solar--hourly}, and \texttt{Bizitobs\_l2c--hourly}. These tasks were chosen solely to maximize diversity in frequency and prediction length; accuracy on this subset is not emphasized, as we continue to use the full set of 97 tasks for performance reporting. For efficiency, we timed each model’s total inference time to generate forecasts on all selected tasks, running on a single H200 GPU and using the official replication code provided in the GIFT-Eval repository.\footnote{Replication code for all leaderboard models is available at: \url{https://github.com/SalesforceAIResearch/gift-eval/tree/main/notebooks}.}

\Cref{fig:perf_size_speed} summarizes the trade-offs between model size, inference time, and predictive performance. Among baseline models, \texttt{Kairos-50M} emerges as the fastest, although it lags behind \modelname{} in benchmark accuracy and is nearly five times larger in parameter count. \texttt{Granite-FlowState-R1}, on the other hand, achieves stronger accuracy than \modelname{} while being smaller in size, but it is roughly three times slower in inference speed. Most notably, \modelname{} demonstrates a substantial leap in efficiency compared to its predecessor: it is approximately 30$\times$ smaller and 2$\times$ faster than Moirai-Large, while simultaneously achieving significantly better accuracy.

\subsection{Scaling Experiments}

We further investigate the effect of scaling model size on performance by training two larger variants of \modelname{}, namely base and large, which increase the parameter count by approximately $8\times$ and $30\times$ compared to the small version. As shown in \Cref{tab:scaling}, we do not observe scaling benefits: both larger variants underperform the small model on GIFT-Eval in terms of MASE and CRPS. In fact, the small variant already appears sufficient to fully exploit the available pretraining data, and simply increasing the parameter count does not translate into better results. This finding suggests that improvements may require further scaling of pretraining data, architectural innovations, or targeted regularization strategies, rather than model size alone.

\begin{wraptable}{r}{0.5\textwidth} \vspace{-2.5 em}
    \centering
    \caption{Performance of different \modelname{} sizes.}
    \label{tab:scaling}%
    \resizebox{0.5\textwidth}{!}{%
    \begin{tabular}{lccc}
    \toprule
    \textbf{Model} & \textbf{Params (M)} & \textbf{MASE} & \textbf{CRPS} \\
    \midrule
    Moirai 2.0 small & 11.4 & \textbf{0.728}  & \textbf{0.516} \\
    Moirai 2.0 base  & 87.1 & 0.732          & 0.525 \\
    Moirai 2.0 large & 305  & 0.743          & 0.530 \\
    \bottomrule
    \end{tabular}%
    }
\end{wraptable}%

This trend is consistent with the broader observations reported in the GIFT-Eval benchmark paper, where scaling across model families does not systematically lead to improved forecasting performance~\citep{aksu2024gift}. At the same time, other recent studies suggest that scaling parameter counts can yield benefits~\citep{feng2025kairos, wang2025output}. We hypothesize that the interaction between model size and pretraining data availability is the critical factor here, without aligned increases in data scale and diversity, parameter growth alone does not guarantee gains~\citep{kaplan2020scaling}.


\subsection{Ablation Study}

We conduct an ablation study to trace the evolution from Moirai 1.0 to Moirai 2.0, with results summarized in \Cref{tab:ablation}. Starting from the original Moirai 1.0 small, which uses an encoder-only architecture and a distributional loss. 

To disentangle the effect of new pretraining data, we first trained a v0 of decoder-only architecture with GIFT-Eval Pretrain, which applies the new decoder-only architecture with the original distributional loss. This resulted in improvements for MASE and minimal change for CRPS. Next in v1, we train the same architecture with the newly curated corpus and observe a substantial improvement on both metrics. The v2 version replaces the distributional loss with quantile loss, yielding the largest single improvement across the ablations. Building on this, v3 augments the decoder-only design with recursive decoding, further reducing error. In v4, we additionally apply random masking during pretraining. While this modification alone slightly degrades performance, we retained it in combination with multi-token prediction, where its benefits became more apparent in v5. Multi-token prediction enabled alongside recursive decoding and random masking yields the best CRPS while retaining similar MASE. The final version of \modelname{} strengthens the projection layer by replacing the linear head with a residual block, while keeping quantile loss, decoder-only architecture, and all three pretraining/inference strategies. This final variant achieves the best overall performance, completing the incremental path of design decisions.

\begin{table}[ht]
\centering
\caption{Ablation study of \modelname{} design choices. Starting from the \prevmodelname{} followed by a decoder-only variant using the original GIFT-Eval pretraining and distributional loss (v0), new pretraining corpus (v1), quantile loss (v2), autoregressive quantile decoding (v3), random masking (v4), and multi-token prediction (v5). The final \modelname{} additionally replaces the linear projection with a residual block, achieving the best overall results.}
\label{tab:ablation}
\begin{adjustbox}{max width=\textwidth}
\begin{tabular}{lccccccccc}
\toprule
\textbf{Variant} & \textbf{ Pretrain Data} & \textbf{Loss} & \textbf{Architecture} & \textbf{Projection} & \textbf{Multitoken pred.} & \textbf{Autoreg. quantiale dec.} & \textbf{Random mask} & \textbf{MASE} & \textbf{CRPS} \\
\midrule
Moirai 1.0 small    &  GIFT-Eval Pretrain & distribution & enc-only      & linear           & \textemdash{} & \textemdash{} & \textemdash{} &  0.946   &  0.65   \\
\midrule
v0  & GIFT-Eval Pretrain & distribution & dec-only      & linear           & \textemdash{} & \textemdash{} & \textemdash{} &  0.9288  &  0.6469  \\
v1      & New Corpus & distribution & dec-only      & linear           & \textemdash{} & \textemdash{} & \textemdash{} & 0.85    & 0.58    \\
v2      & New Corpus & quantile     & dec-only      & linear           & \textemdash{} & \textemdash{} & \textemdash{} & 0.744 & 0.553 \\
v3      & New Corpus & quantile     & dec-only  & linear           & \textemdash{} & \checkmark & \textemdash{} & 0.736 & 0.533  \\
v4      & New Corpus & quantile     & dec-only  & linear           & \textemdash{} & \checkmark            & \checkmark & 0.772 & 0.56 \\
v5      & New Corpus & quantile     & dec-only  & linear           & \checkmark             & \checkmark             & \checkmark             & 0.739 & 0.527 \\
\midrule
\modelname{} & New Corpus & quantile     & dec-only  & residual block   & \checkmark             & \checkmark             & \checkmark             & 0.728   & 0.516   \\
\bottomrule
\end{tabular}
\end{adjustbox}
\end{table}















\section{Limitations and Future Work}
\label{sec:fworkandlimitations}

Seeing minimal benefit in doing so, we have dropped support for multivariate forecasting and the use of covariates in \modelname{}. This observation likely stems from the limited availability of high-quality datasets with such properties, which may be revisited in future versions once this limitation is resolved — potentially through synthetic data generation, as demonstrated by~\citet{ansari2025chronos2}.


While \modelname{} benefits substantially from its architectural changes—achieving better performance with a smaller parameter count—it also introduces new challenges for scaling. Counterintuitively, increasing the number of parameters degrades performance, suggesting a mismatch between the model architecture and the available data. Furthermore, the model exhibits weaker performance as the forecast horizons gets longer. In future work we would like to explore bridging this gap between architecture and data, enabling scalable model growth and potentially yielding further performance improvements. 

For future work there are several exciting directions we believe are worth to explore. First, agentic solutions that combine time-series analysis capabilities with the reasoning power of large language models (LLMs) represent a highly promising avenue. The recent agent-based submissions to the GIFT-Eval benchmark further demonstrate the potential of this paradigm. We anticipate that such approaches will become increasingly practical and relevant as new benchmarks emerge that demand deeper reasoning and contextual understanding.

Another important direction lies in developing foundation models capable of multimodal reasoning, integrating text, image, and time-series modalities. While some early works have begun exploring this space~\citep{cai_jolt_2024, niu_langtime_2025,wang_chattime_2024,xie_chatts_2024,chow_towards_2024,merrill_language_2024}, much remains to be discovered. We believe that forecasting and time-series analysis will become far more valuable for both businesses and everyday users when enriched with additional contextual information which makes multimodal integration an exciting path for future research.


\section{Conclusion}
\label{sec:conclusion}
In this technical report we introduced our new model, \modelname{} a decoder-only time-series foundation model that pairs quantile forecasting with multi-token prediction and trained on a newly curated diverse pretraining data with 36 million time series. Evaluated on the comprehensive \textsc{Gift-Eval} benchmark, \modelname{} is highly competitive—achieving top 5 performance among pretrained models while offering a favorable trade-off between accuracy, inference speed, and parameter count.
Our ablations isolate the effect of each design choice and show that switching from encoder-only to decoder-only architecture, together with recursive multi-quantile decoding accounts for most of the observed gains.

Along these advances, we also discussed limitations of \modelname{}. We observed diminishing returns from parameter scaling, and degraded performance as forecast horizons lengthen. Addressing these challenges may require: data scaling aligned with model capacity; and architectural improvements aimed at long-horizon performance. By releasing code, documenting and evaluating design choices, and reporting limitations, we hope to support and accelerate progress in this direction.



\clearpage
\newpage

\bibliography{references}
\bibliographystyle{plainnat}

\clearpage
\newpage

\appendix
\section{Appendix}

\begin{figure}[!htb]
    \centering
    \includegraphics[width=1\linewidth]{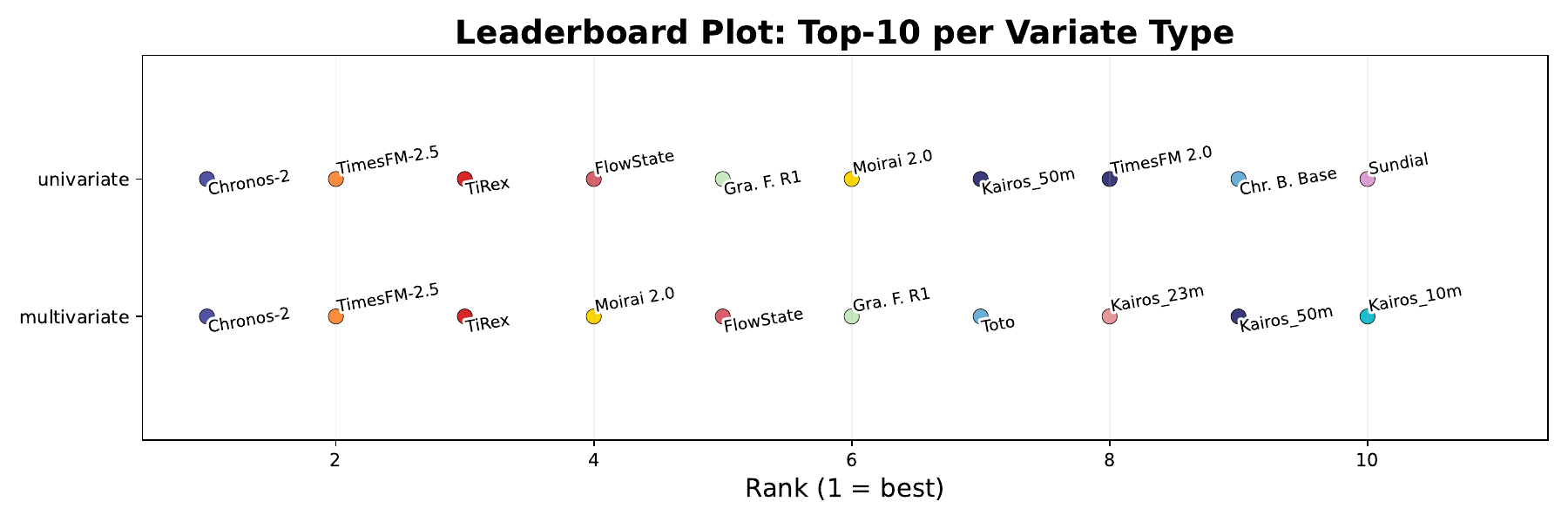}
    \caption{\textsc{GiftEval} leaderboard results broken down by variate type. For both univariate and multivariate, we display the top-10 foundation models ordered by their MASE rank (lower is better).}
    \label{fig:leaderboard-per-variate}
\end{figure}

\begin{figure}[!htb]
    \centering
    \includegraphics[width=1\linewidth]{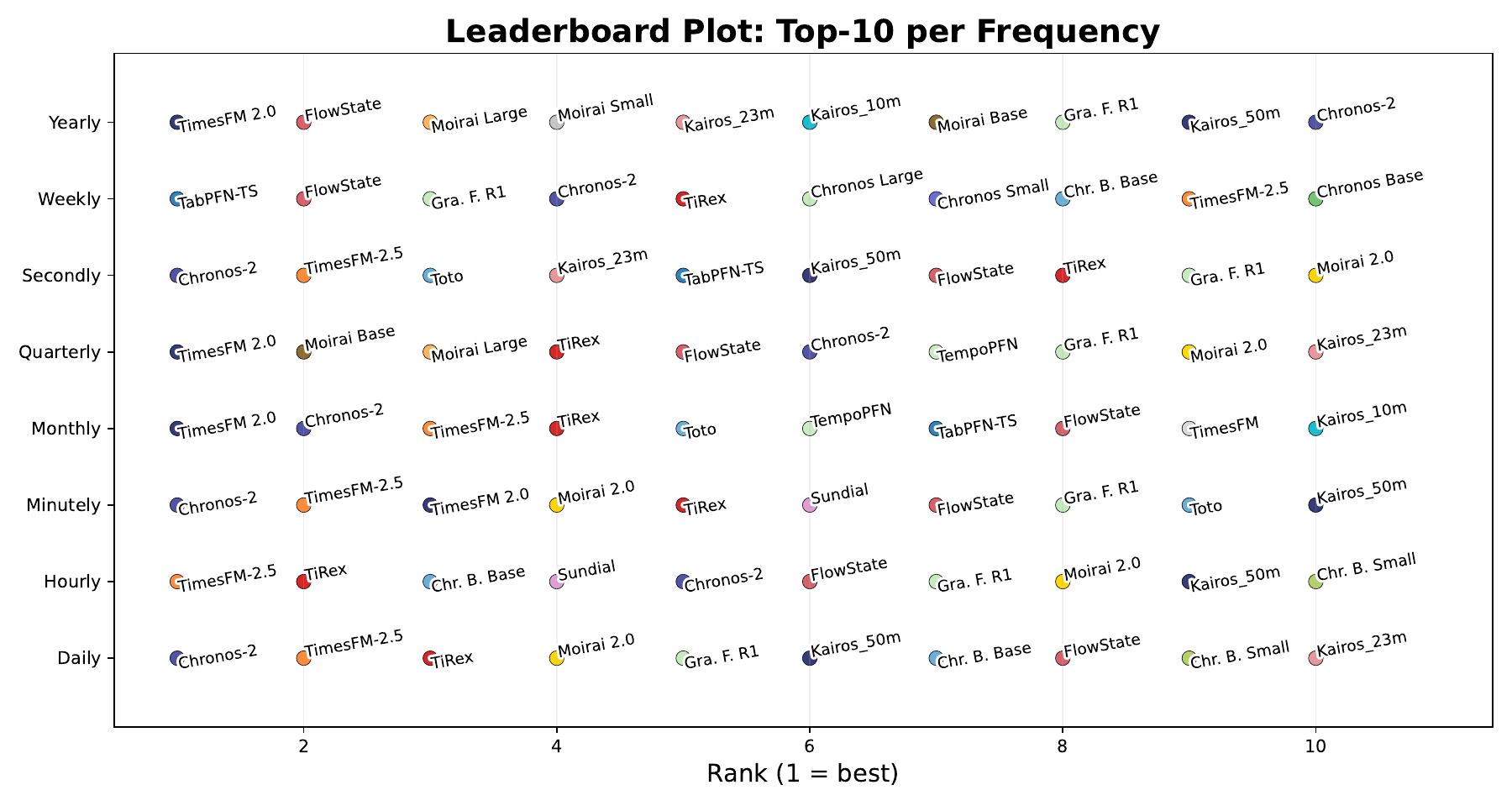}
    \caption{\textsc{GiftEval} leaderboard results broken down by frequency. For each frequency type, we display the top-10 foundation models ordered by their MASE rank (lower is better).}
    \label{fig:leaderboard-per-freq}
\end{figure}

\begin{figure}[!htb]
    \centering
    \includegraphics[width=1\linewidth]{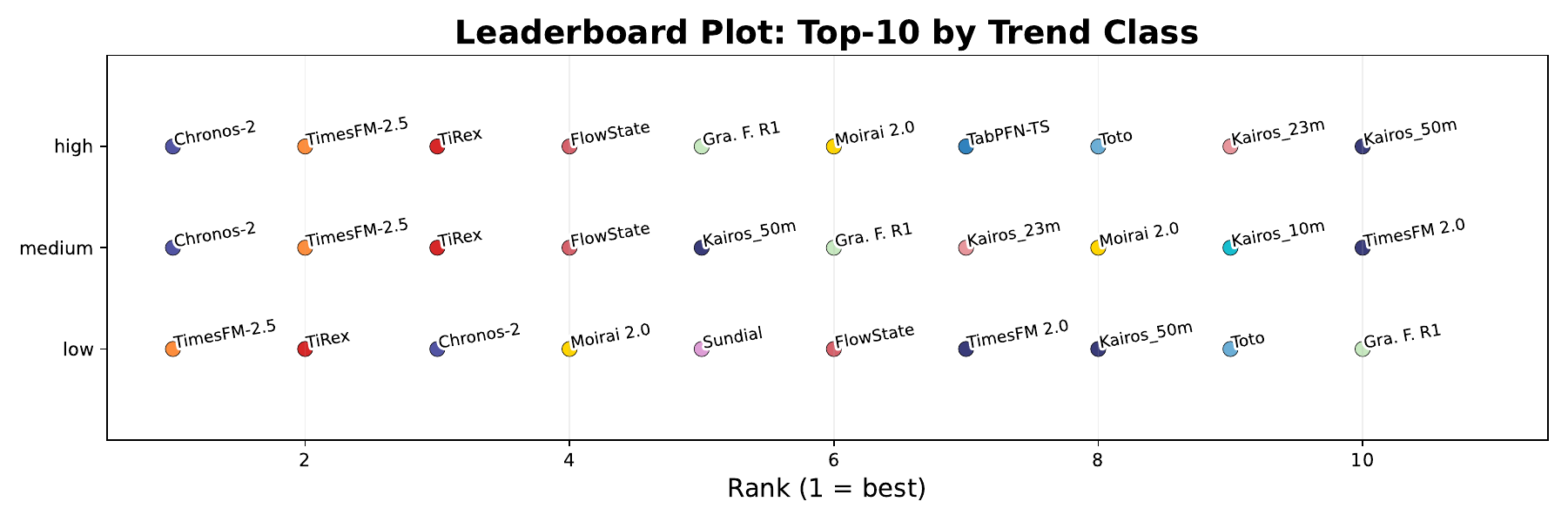}
    \caption{\textsc{GiftEval} leaderboard results broken down by trendiness of time series. High value indicates strong trend and vice versa.  We display the top-10 foundation models ordered by their MASE rank (lower is better).}
    \label{fig:leaderboard-per-trend}
\end{figure}

\begin{figure}[!htb]
    \centering
    \includegraphics[width=1\linewidth]{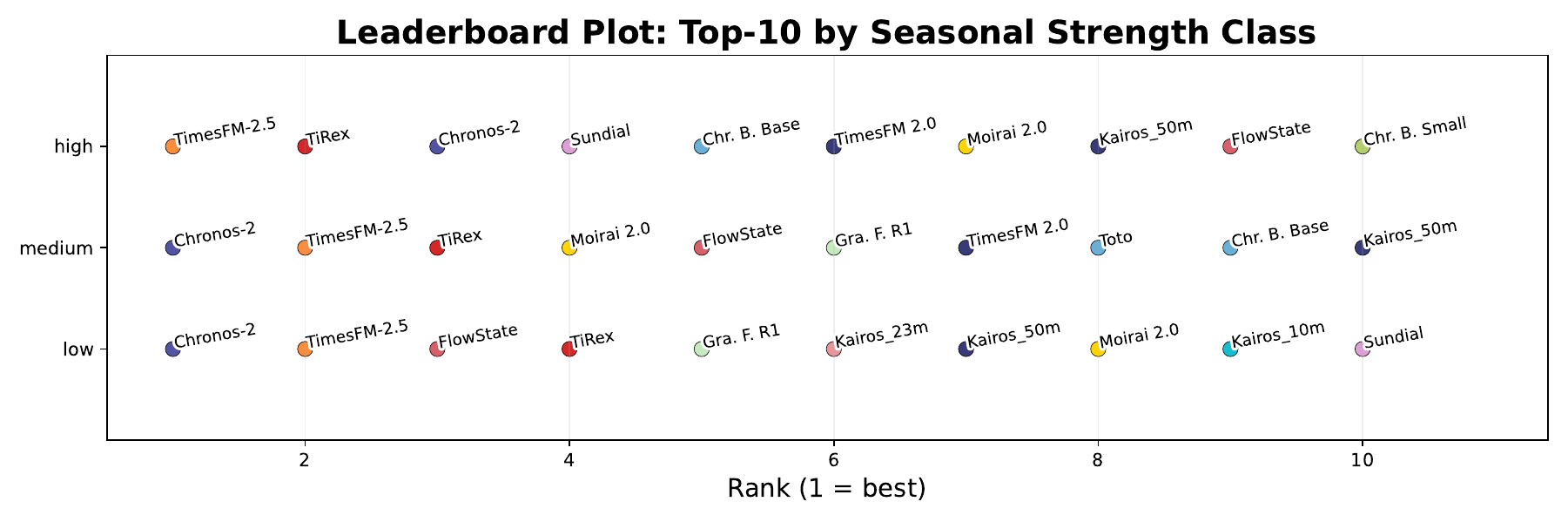}
    \caption{\textsc{GiftEval} leaderboard results broken down by seasonal strength of the time series. High value indicates strong seasonal patterns and vice versa.  We display the top-10 foundation models ordered by their MASE rank (lower is better).}
    \label{fig:leaderboard-per-seas-str}
\end{figure}

\begin{figure}[!htb]
    \centering
    \includegraphics[width=1\linewidth]{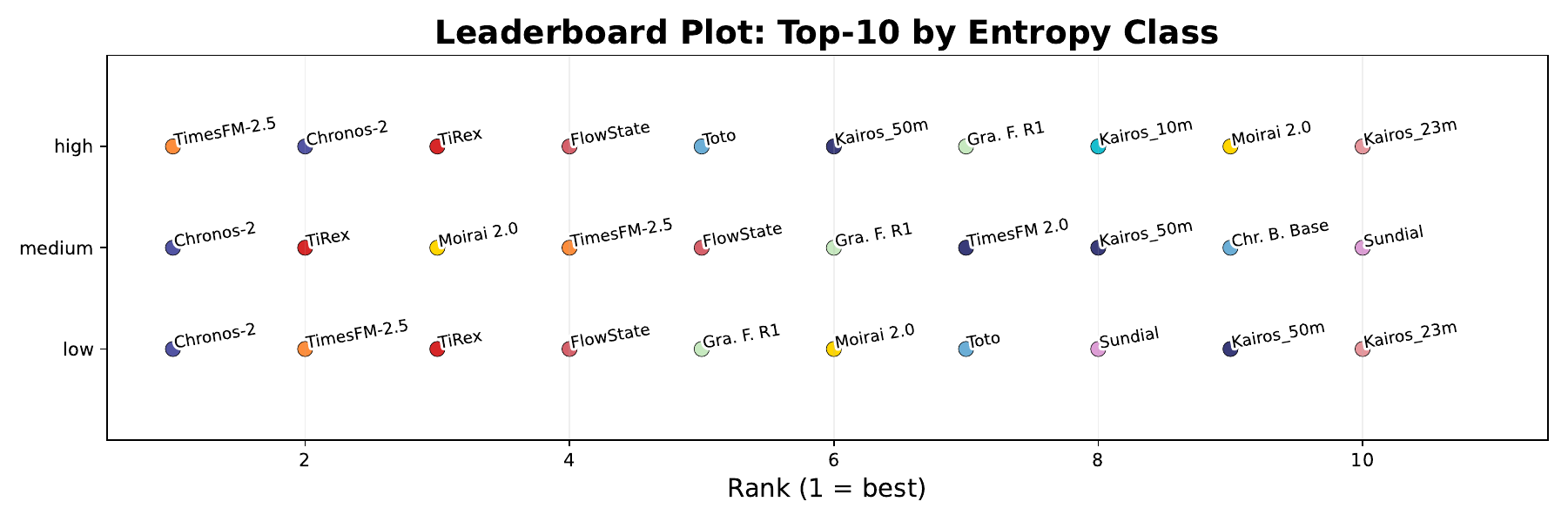}
    \caption{\textsc{GiftEval} leaderboard results broken down by entropy of the time series. Entopy measures the “forecastability” of a time series, where low values indicate a high signal-to-noise ratio and high values occur when a series is difficult to forecast.  We display the top-10 foundation models ordered by their MASE rank (lower is better).}
    \label{fig:leaderboard-per-entropy}
\end{figure}

\begin{figure}[!htb]
    \centering
    \includegraphics[width=1\linewidth]{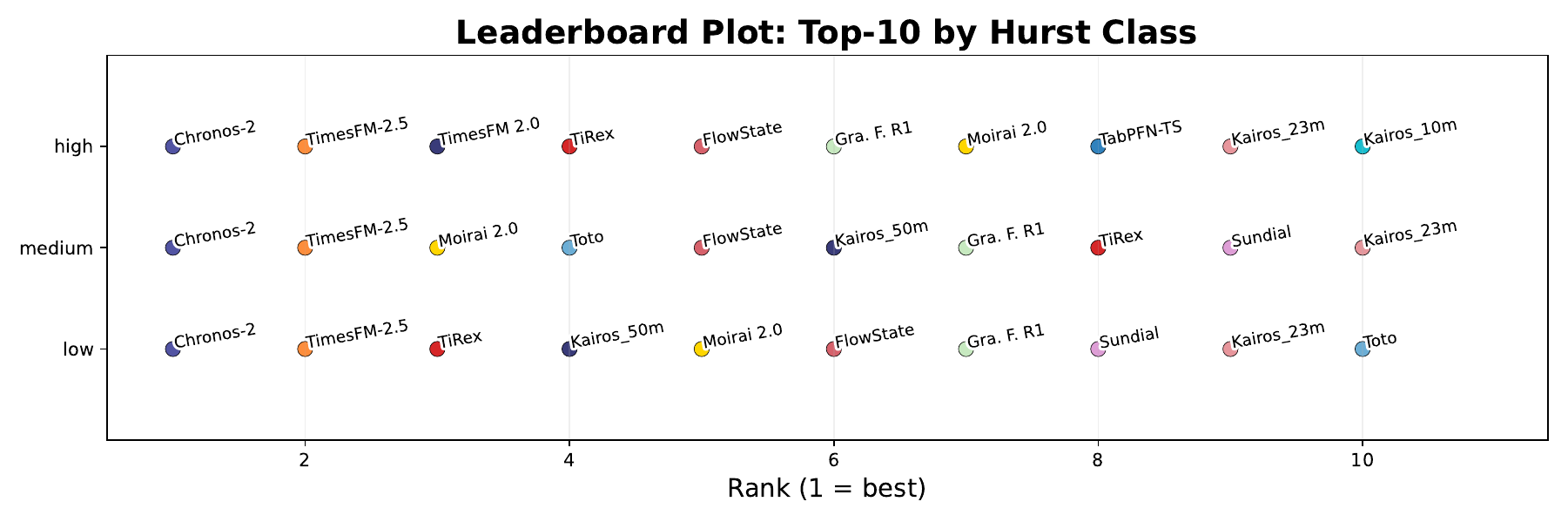}
    \caption{\textsc{GiftEval} leaderboard results broken down by hurts value of the time series.  Hurst indicates the long-term memory or persistence of a time series and whether future values are likely to be influenced by past trends, revert to the mean, or behave randomly. Higher values indicate more persistence.  We display the top-10 foundation models ordered by their MASE rank (lower is better).}
    \label{fig:leaderboard-per-hurst}
\end{figure}

\begin{figure}[!htb]
    \centering
    \includegraphics[width=1\linewidth]{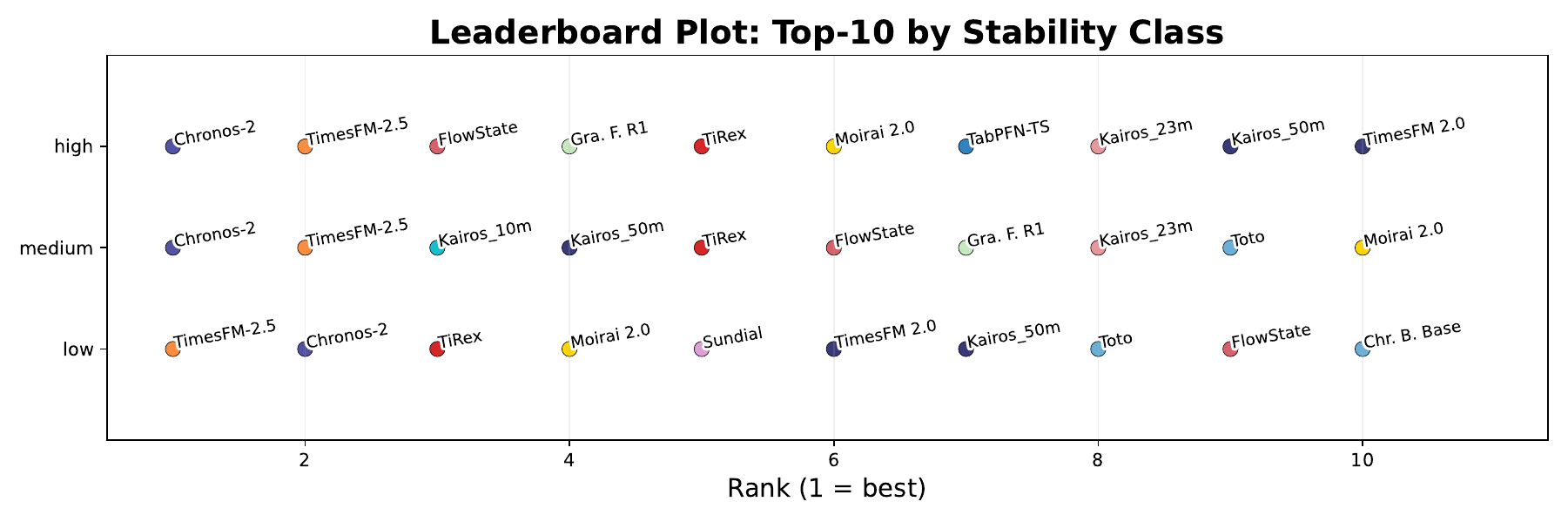}
    \caption{\textsc{GiftEval} leaderboard results broken down by stability of the time series.   Stability is the variance of the means. Lower values indicate more stable data. We display the top-10 foundation models ordered by their MASE rank (lower is better).}
    \label{fig:leaderboard-per-stability}
\end{figure}

\begin{figure}[!htb]
    \centering
    \includegraphics[width=1\linewidth]{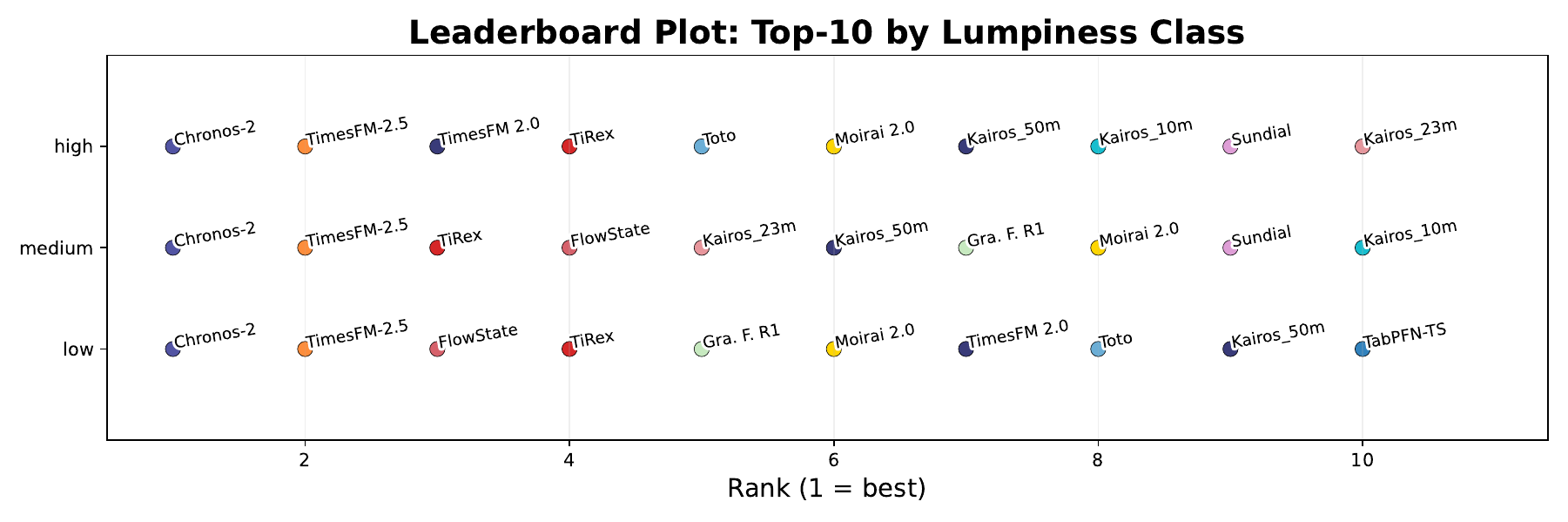}
    \caption{\textsc{GiftEval} leaderboard results broken down by lumpiness of the time series. Lumpiness is the variability of the variance across different segments of the time series. A high value of lumpiness indicates significant fluctuations in variability. We display the top-10 foundation models ordered by their MASE rank (lower is better).}
    \label{fig:leaderboard-per-lumpiness}
\end{figure}

\end{document}